\documentclass{article}

\usepackage{PRIMEarxiv}

\usepackage[utf8]{inputenc} 
\usepackage[T1]{fontenc}    
\usepackage{hyperref}       
\usepackage{url}            
\usepackage{booktabs}       
\usepackage{amsfonts}       
\usepackage{nicefrac}       
\usepackage{microtype}      
\usepackage{fancyhdr}       
\usepackage{graphicx}       
\usepackage{natbib}
\usepackage{float}
\graphicspath{{media/}}     
\setcitestyle{authoryear,open={(},close={)}}

\title{Impact of Domain-Adapted Multilingual Neural Machine Translation in the Medical Domain
\thanks{\textit{\underline{Citation}}: 
\textbf{Work presented at Translating and the Computer - TC44, 24 November 2022, Luxembourg. \url{https://asling.org/tc44/confirmed-presentations/\#Rios-Gaona}}} 
}

\author{
  Miguel Rios, Raluca-Maria Chereji, Alina Secară, Dragoș Ciobanu \\
  Centre for Translation Studies, University of Vienna \\
  \texttt{\{miguel.angel.rios.gaona, raluca-maria.chereji,} \\ \texttt{alina.secara, dragos.ioan.ciobanu\}@univie.ac.at} \\
}

\begin{document}
\maketitle

\begin{abstract}

Multilingual Neural Machine Translation (MNMT) models leverage many language pairs during training to improve translation quality for low-resource languages by transferring knowledge from high-resource languages. We study the quality of a domain-adapted MNMT model in the medical domain for English-Romanian with automatic metrics and a human error typology annotation which includes terminology-specific error categories. We compare the out-of-domain MNMT with the in-domain adapted MNMT. The in-domain MNMT model outperforms the out-of-domain MNMT in all measured automatic metrics and produces fewer terminology errors.

\end{abstract}


\section{Introduction}

Current state-of-the-art Neural Machine Translation (NMT) models have shown promising results on low-resource language pairs, particularly for non-specialised domains \citep{araabi_optimizing_2020}. However, in a high-risk and low-resource domain, like the medical domain, the accurate translation of terminology is crucial for exchanging information across international healthcare providers or users \citep{skianis_evaluation_2020}. Multilingual NMT (MNMT) models leverage many language pairs and millions of segments during training \citep{johnson_googles_2017}. The inclusion of many language pairs helps to improve the translation quality for low-resource languages by transferring knowledge from high-resource languages. Moreover, domain adaptation techniques are used to adapt MNMT models into new domains \citep{berard_multilingual_2020}. However, evaluation studies of MNMT models are focused on automatic metrics without providing insights into the quality of the translation of specialised terminology.

In this paper, we study the quality of a pre-trained MNMT model in the medical domain for a low-resource language pair (English-Romanian). Our goal is to compare an out-of-domain MNMT with a fine-tuned in-domain MNMT in terms of automatic metrics and terminology translation. We use a pre-trained model based on MBart \citep{liu_multilingual_2020} and fine-tune it with a medical in-domain parallel corpus.

We test the models on the English-Romanian language pair with a corpus of medical paper abstracts 
\citep{neves_findings_2018}. We evaluate both models with automatic metrics, and a terminology error typology annotation performed by in-house human annotators \citep{haque_investigating_2019}. The fine-tuned MBart model outperforms MBart on the automatic metrics. In addition, the error analysis based on a terminology-based error typology \citep{haque_investigating_2019} shows that the fine-tuned model also produces fewer errors than the MBart model.

\section{Background}

MNMT models are based on transferring parameters or information across multiple languages, where low-resource languages benefit from the high-resource languages. The MNMT model shares a common word representation (i.e., arrays of numbers) across language pairs. During training, the MNMT model clusters words with similar contexts from the high- and low-resource segments \citep{johnson_googles_2017}. The low-resource pairs learn meaningful word representations given the access to a large number of similar contexts from the high-resource language pairs. Moreover, an MNMT model allows to translate across multiple languages by using only one translation system. The multiple languages are processed jointly by indicating the target translation direction on each segment of the multilingual corpora in the input training data by using an artificial token (label \textit{<2target>}). For example, an English-Romanian segment pair would be labelled as follows:

\textit{<2ro> It is noted that in some cases increase of blood pressure was documented. $\rightarrow$ Se remarcă faptul că, în unele cazuri, s-a înregistrat creșterea tensiunii arteriale.
}

MNMT models outperform standard bilingual baselines on translation quality for low-resource languages \citep{johnson_googles_2017}, but they require a high amount of computational resources to process the millions of parallel multilingual segments.

In particular, MBart is a sequence-to-sequence model pre-trained on monolingual data from 25 languages based on a text reconstruction learning objective for MNMT \citep{liu_multilingual_2020}. MBart incorporates a monolingual training step before the multilingual MT training for a better initialisation of the translation model. In other words, MBart first learns an improved individual representation of each language with monolingual data. After that, MBart continues with the multilingual translation training based on parallel data. MBart shows a better translation quality compared to previous MNMT models.

However, most MNMT models are general-purpose systems trained with web crawled corpora \citep{liu_multilingual_2020, verma_strategies_2022}, and they struggle with specialised domains (e.g. medical). Domain adaptation aims to improve the translation performance in specialised domains, where fine-tuning is a low-cost and common technique. Fine-tuning consists of resuming the training of an out-of-domain resource-rich MT model with a poor-resourced in-domain corpus \citep{chu_survey_2018}. The resulting model is adapted to work with an in-domain language pair, instead of re-training the MNMT model from scratch \citep{verma_strategies_2022}.

MT models are usually evaluated with automatic metrics that take into account fluency and adequacy, by comparing the machine translation output against one or more human reference translations \citep{papineni_bleu_2002}. Metrics produce a corpus-level score or a segment-level score for a given MT model \citep{rei_comet_2020}. However, automatic metrics are not designed to identify translation errors in MT outputs, for example, errors in terminology \citep{haque_investigating_2019}. On the other hand, error typology evaluation frameworks, such as the Multidimensional Quality Metrics (MQM) \citep{lommel_multidimensional_2013}, are based on manually classifying and annotating errors using predefined categories. \citet{haque_investigating_2019} propose an error typology with a focus on terminology: human evaluators identify an error in the MT output, select a category out of the eight available, and assign a severity score.

\section{Experiments}

For fine-tuning, we use the English-Romanian section from the EMEA parallel corpus \citep{clarinel_emea_2015}. The EMEA corpus consists of PDF documents from the European Medicines Agency. We split the corpus into $775,904$ training, and $7,837$ validation segments. We evaluate the MNMT models with the test dataset of abstracts from scientific publications from Medline \citep{neves_findings_2018} which contains $291$ segments.

We use BLEU \citep{papineni_bleu_2002,post_call_2018}, chrF \citep{popovic_chrf_2015}, and COMET \citep{rei_comet_2020} for automatic evaluation. For human evaluation, we use \citep{haque_investigating_2019} which contains eight terminology-related error categories - Partial error, Source term copied, Inflectional error, Reorder error, Disambiguation issue in target, Incorrect lexical selection, Term drop, and Other error -, and three severity levels - Minor, Major and Critical.

We continue training MBart with the EMEA corpus to adapt it into the medical domain, and we perform model selection using BLEU on the validation split. We performed our experiments with Fairseq \citep{ott_fairseq_2019} using an open-source pre-trained model for MBart\footnote{\url{https://dl.fbaipublicfiles.com/fairseq/models/mbart/mbart.cc25.ft.enro.tar.gz}}. The settings for the fine-tuned MBart are as follows: Adam with learning rate $3\mathrm{e}{-5}$, inverse square root scheduler, $2,500$ warm-up updates, $40,000$ updates, dropout $0.3$, attention dropout $0.1$, label smoothing $0.2$, batch size $2048$ tokens ($256$ maximum tokens per batch, and $8$ batches for gradient accumulation), and memory efficient fp16 training. We used a 16GB Tesla T4 GPU from the Google Cloud platform for training\footnote{The scripts for our experiments are available at: \url{https://github.com/mriosb08/medical-NMT-HAITrans}}. The fine-tuning process took 38 hours to complete.

\subsection{Results}

We define general MBart (out-of-domain data), and fine-tuned MBart (in-domain medical data). Table \ref{tab:metrics} shows the automatic metrics scores for both models. Fine-tuned MBart outperforms the general model on all the metrics. The BLEU score is statistically significant $p=0.001$ based on bootstrap resampling with $1,000$ iterations.

\begin{table}[]
\centering
\begin{tabular}{lccc}
\toprule
                 & BLEU $\uparrow$ & chrF $\uparrow$ & COMET $\uparrow$ \\ \hline
MBart            & 21.9   & 51.5   & 0.556   \\
fine-tuned MBart & \textbf{25.8}   & \textbf{54.9}   & \textbf{0.663}   \\ \bottomrule
\end{tabular}
\caption{Automatic metrics for MBart and fine-tuned MBart.}
\label{tab:metrics}
\end{table}

\begin{figure}[h]
\centering
\includegraphics[width=12.5cm]{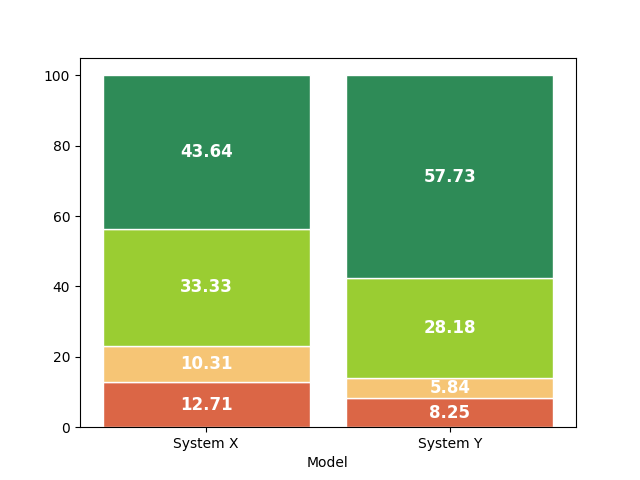}
\caption{COMET scores for segments divided into quality bins for MBart (System X) and fine-tuned MBart (System Y).}
\label{fig:comet1}
\end{figure}

Furthermore, we performed an analysis of the COMET segment level scores. We use MT-Telescope \citep{rei_mt-telescope_2021} to compare both systems. Figure \ref{fig:comet1} shows the percentage of segments divided into four quality bins. Each bin is defined by a default threshold from the COMET scores, from green (residual errors) to red (critical errors). The red bin has translations lower than $0.10$ score, the yellow bin has translations between $0.10$ and $0.30$ score, the light green has translations between $0.30$ and $0.70$ score, and the dark green has translations greater than $0.70$ score. MBart is System X and fine-tuned MBart is System Y. The fine-tuned MBart has the highest number of high scores compared to the original general model.

\begin{figure}[h]
\centering
\includegraphics[width=17cm]{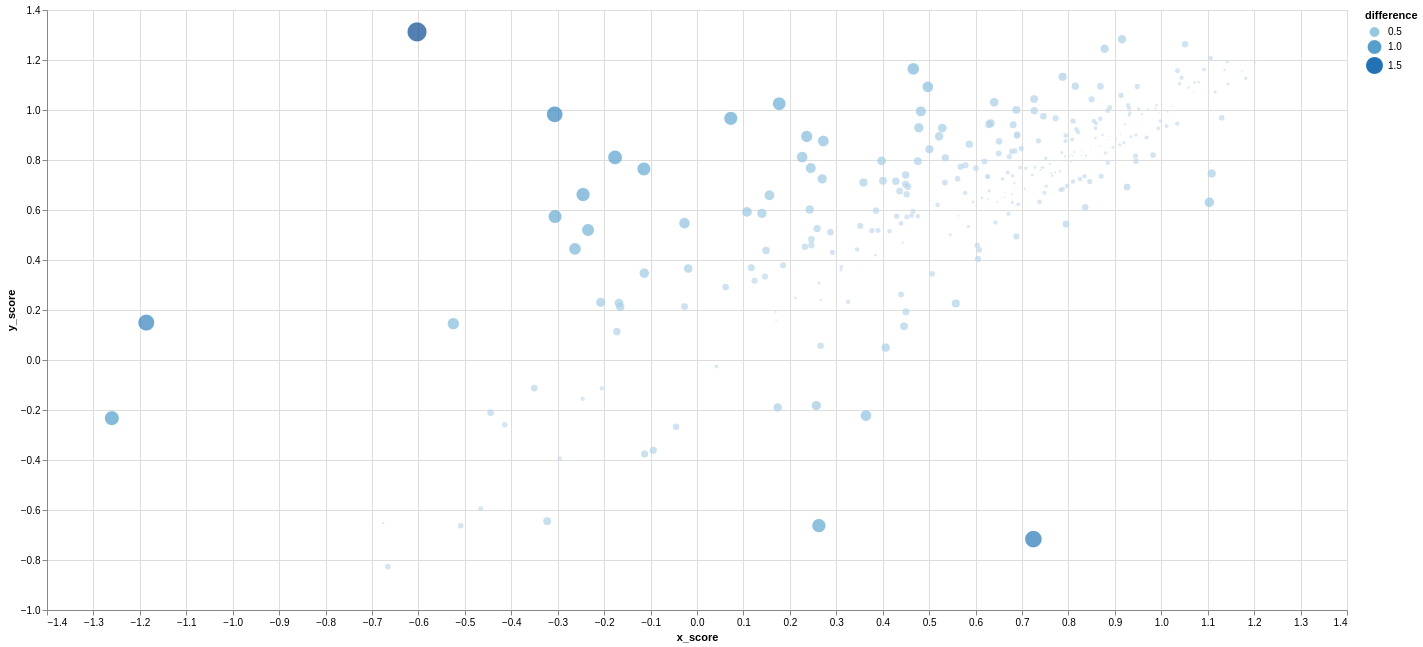}
\caption{Bubble plot of segment level COMET scores for MBart (x\_score) and fine-tuned MBart (y\_score).}
\label{fig:comet2}
\end{figure}

Figure \ref{fig:comet2} shows visually the difference of COMET scores between the two systems for each segment. The size and colour of a bubble is the difference in the COMET score between systems for the same segment. Moreover, systems are different when the bubbles are far from each other along the axis (x\_score MBart, and y\_score fine-tuned MBart), and from the centre of the plot. Both systems MBart and fine-tuned MBart are different in terms of COMET scores, and fine-tuned MBart has a higher COMET score. If both models produce different translations, in this case, it means that the fine-tuned model is learning to generate MT outputs close to the medical domain.

\subsection{Error Analysis}

To gain insights into the specific terminology errors produced by the two models, we show a sample of $12$ abstracts with a total of $75$ segments to three annotators. The annotators are native Romanian speakers with in-house and freelance translation experience; moreover, one of the annotators also has in-house and freelance medical translation experience. The annotators had access to the source, the reference, and the output of the two MT systems to annotate each MT segment with error categories \citep{klubicka_fine-grained_2017} using \citep{haque_investigating_2019}. The annotators annotated the abstracts collaboratively \citep{esperanca-rodier_accole_2019} – the motivation for the joint in-person annotation is to increase agreement for identifying possible terms and errors.

To perform the annotation, we set up a translation project in Trados Studio $2021$\footnote{\url{https://www.trados.com/products/trados-studio/}} and import the source, reference and MT output files as bilingual .xlsx files. We install the freely-available Qualitivity \footnote{\url{https://community.rws.com/product-groups/trados-portfolio/rws-appstore/w/wiki/2251/qualitivity}} plugin integrated into Studio using an API key; this serves as the environment in which the annotators record any identified errors, their severity level and proposed corrections, along with explanatory comments. At the end of the annotation process, we export a report from Qualitivity containing the full annotation data for the reference texts, and MBart and fine-tuned MBart outputs.

The total number of terminology-related errors for general-model MBart and fine-tuned MBart are $98$ and $64$ respectively, demonstrating the improvement brought about by the fine-tuning process with in-domain data. Table \ref{tab:mqmerr} shows the number of errors for each category present in the abstracts.

\begin{table}[]
\centering
\begin{tabular}{lcc}
\toprule
Error Type                     & MBart $\downarrow$ & fine-tuned    MBart $\downarrow$ \\ \hline
Partial error                  & 41      & \textbf{23}                    \\
Source term copied             & 22      & \textbf{9}                     \\
Inflectional error             & \textbf{2}       & 4                     \\
Reorder error                  & \textbf{1}       & 3                     \\
Disambiguation issue in target & 14      & \textbf{6}                     \\
Incorrect lexical selection    & 9       & \textbf{6}                     \\
Term drop                      & 0       & 0                     \\
Other error                    & \textbf{9}       & 13                    \\ 
\bottomrule
\end{tabular}
\caption{Total errors for each terminology-related category.}
\label{tab:mqmerr}
\end{table}

The fine-tuned MBart model produces fewer errors than the general model on most of the categories. However, the fine-tuned model fails in the following categories: Inflectional error, Reorder error, and Other. Moreover, we show annotated examples of random segments for each error category to further understand the cause of the errors. In Table \ref{tab:err1} we show a random selection of source and MT output for each error category, except Other, and highlight the annotated errors for each category for fine-tuned MBart.

\begin{table}[t]
\centering
\begin{tabular}{p{0.15\textwidth}p{0.35\textwidth}p{0.35\textwidth}}
\toprule
\multicolumn{1}{c}{Category}     & \multicolumn{1}{c}{Source}                                                                                                                                                                                                                                                                                           & \multicolumn{1}{c}{Target (fine-tuned   MBart)}                                                                                                                                                                                                                                                                                                               \\ \hline
Partial error                    & The DX-OSA score may be   useful for identifying obese patients with significant OSA who require CPAP   (continuous positive airway pressure) treatment, and CPAP could be commenced   without the need for polysomnography, therefore, without delaying surgery.                                                    & Scorul DX-OSA poate fi   util pentru identificarea pacienţilor obezi cu OSA semnificativă care   necesită tratament cu CPAP (\textbf{tensiune   arterială continuă pozitivă} {[}instead   of \textbf{presiune pozitivă continuă în căile aeriene}{]}), iar CPAP poate fi   început fără a fi necesară polisomnografie, prin urmare, fără a întârzia   intervenţia chirurgicală. \\ \hline
Source term copied               & The objectives of this study were   to reveal possible relations between antioxidant therapy and a number of   serum biochemical variables (ALT, AST, APPT, LDH, urea, leukocytes,   platelets), the length of mechanical ventilation, the time spent in the ICU,   and the mortality rate in major trauma patients. & Obiectivul acestui studiu a fost să   evidențieze posibilele relații dintre tratamentul cu antioxidanti și o serie   de variabile biochimice serice (ALT, AST, \textbf{APPT} {[}instead of \textbf{APTT}{]}, LDH, uree, leucocite, trombocite), durata ventilației mecanice, timpul   petrecut în ICU și rata mortalității la pacienții cu traumatisme majore.                  \\ \hline
Inflectional error               & Two of these drugs, duloxetine and   venlafaxine, are used also in chronic pain management.                                                                                                                                                                                                                          & Două dintre aceste medicamente, \textbf{duloxetină} şi \textbf{venlafaxină} {[}instead of \textbf{duloxetina} și \textbf{venlafaxina}{]}, sunt utilizate şi în tratamentul durerii cronice.                                                                                                                                                                                                       \\ \hline
Reorder error                    & Although not statistically   significant, MODS and ARDS incidences were higher in the DCO shock group:   MODS (41.7\% versus 22.6\% and 20\%; p = 0.08/0.17), ARDS (29.2\% versus 17\% and   20\%; p = 0.22/0.53).                                                                                                   & Deşi nu au fost semnificative   statistic, incidenţele MODS şi ARDS au fost mai mari în \textbf{grupul cu şoc DCO} {[}instead of \textbf{grupul DCO cu șoc}{]}: MODS (41,7\% faţă de   22,6\% şi 20\%; p = 0,08/0,07), ARDS (29,2\% faţă de 17\% şi 20\%; p = 0,22/0,53).                                                                                                       \\ \hline
Disambiguation issue in   target & The drug's efficacy results from   its modulating effect on the descending inhibitory pain pathways and the   inhibition of the nociceptive input.                                                                                                                                                                   & Eficacitatea medicamentului rezultă   din efectul său de modulare asupra \textbf{căilor de durere inhibatoare descendente}   {[}instead of \textbf{căilor descendente inhibitorii ale durerii}{]} și inhibarea   contribuției nociceptive.                                                                                                                                      \\ \hline
Incorrect lexical   selection    & These results correlate with a   higher trauma score in these patients, more serious lesions requiring several   damage control procedures.                                                                                                                                                                          & Aceste   rezultate sunt corelate cu un \textbf{scor traumatic} {[}instead of \textbf{gravitatea traumatismelor}{]} mai mare la   acești pacienți, leziunile mai grave necesitând mai multe proceduri de   control al leziunilor.                                                                                                                                                \\ \bottomrule

\end{tabular}
\caption{Fine-tuned MBart annotated examples for each error category (except Other error).}
\label{tab:err1}
\end{table}

Table \ref{tab:err2} shows all the examples for the Other error category for the fine-tuned MBart. As the fine-tuned model underperformed in terms of Other errors - $13$ to $9$ -, we investigate this further and list all the annotated errors within the Other category in Table \ref{tab:err2}. We identify two phenomena regarding the treatment of English borrowings and acronyms, and evidence of hallucination. The first phenomenon observed is that source terms are translated, even where a borrowing from English would be the correct translation strategy. For example, \textit{Early Total Care and Damage Control Orthopaedics} lead to translations based on erroneous lexical selection: \textit{metode de control al daunelor}, and \textit{principii de îngrijire în primii ani de viaţă}, respectively, instead of retaining the original source terms in English. Moreover, for \textit{burst} and \textit{burst (suppression)}, the fine-tuned model produces the translations \textit{arsură} and \textit{(supresie) pulmonară} belonging to the lexical fields of burn and bust, pointing to challenges with the setup of the Byte pair encoding (BPE) vocabulary in NMT \citep{araabi_how_2022,lignos_challenges_2019}. Secondly, when acronyms should have been maintained as per the EN source, for instance \textit{MODS}, \textit{DCO}, \textit{ARDS}, and \textit{OS}, they are instead randomly recomposed as \textit{SMO}, \textit{COD}, \textit{SRA}, and \textit{SSO}. Acronyms corresponding to terms with a translation into Romanian are also randomly recomposed, for example \textit{FR} is translated as \textit{RF} rather than \textit{RL}. Finally, there is also an example of a hallucination, the English \textit{intramedullary (nailing)} is erroneously translated by adding a Romanian inflection at the end: \textit{(nailing) intramedullar}, instead of \textit{tijă centromedulară}.

\begin{table}[H]
\centering
\begin{tabular}{p{0.45\textwidth}p{0.45\textwidth}}
\toprule
\multicolumn{1}{c}{Source}                                                                                                                                                                                                                                                                                                               & \multicolumn{1}{c}{Target (fine-tuned   MBart)}                                                                                                                                                                                                                                                                                                                          \\ \hline
The aim of this study   was to evaluate the frontal intracortical connectivity during deep   anaesthesia (burst-suppression).                                                                                                                                                                                                            & Scopul acestui studiu a   fost să evalueze conectivitatea intracorticală frontală în timpul anesteziei   profunde (\textbf{supresie   pulmonară}).                                                                                                                                                                                                                                \\ \hline
Rats were maintained in   deep level anaesthesia (burst-suppression).                                                                                                                                                                                                                                                                    & Ratii s-au menţinut în   anestezie profundă (\textbf{supresie   pulmonară}).                                                                                                                                                                                                                                                                                                      \\ \hline
The global cortical   connectivity (0.5-100 Hz) was 0.61 ± 0.078 during the burst periods compared   to 0.55 ± 0.032.                                                                                                                                                                                                                    & Conectivitatea corticală   globală (0,5-100 Hz) a fost de 0,61 ± 0,078 în timpul perioadelor de \textbf{arsură} comparativ cu 0,55 ±   0,032.                                                                                                                                                                                                                                     \\ \hline
The global cortical   connectivity increased during the burst periods.                                                                                                                                                                                                                                                                   & Conectivitatea corticală   globală a crescut în timpul perioadelor de \textbf{arsură}.                                                                                                                                                                                                                                                                                            \\ \hline
Once the "two event   model" was accepted, it became clear that patients although initially   resuscitated, but in a vulnerable condition, have a high risk that a   secondary aggression (for example, surgical interventions) would precipitate   a state of hyperinflammation and early multiple organ dysfunction syndrome   (MODS). & Odată ce „modelul celor   două evenimente” a fost acceptat, a devenit clar că pacienţii, deşi iniţial   resuscitaţi, dar aflaţi într-o stare vulnerabilă, prezintă un risc crescut ca   o agresivitate secundară (de exemplu intervenţii chirurgicale) să precipite o   stare de hiper inflamaţie şi sindrom de disfuncţie multiplă precoce (\textbf{SMO}).                       \\ \hline
This is a retrospective   study of severe polytrauma patients with femoral shaft fractures admitted to   the intensive care unit of the Emergency clinical Hospital of Bucharest and   treated from an orthopaedic point of view by either Damage Control   Orthopaedics (DCO) or Early Total Care (ETC) principles.                     & Acesta este un studiu   retrospectiv la pacienţi cu politraum sever, cu fracturi ale căilor femurale,   internaţi în unitatea de terapie intensivă a Spitalului clinic de urgenţă din   Bucureşti şi trataţi din punct de vedere ortopedic, fie conform principiilor   de \textbf{control al   deteriorării (DCO)}, fie conform   principiilor de îngrijire totală precoce (ETC). \\ \hline
This is a retrospective   study of severe polytrauma patients with femoral shaft fractures admitted to   the intensive care unit of the Emergency clinical Hospital of Bucharest and   treated from an orthopaedic point of view by either Damage Control   Orthopaedics (DCO) or Early Total Care (ETC) principles.                     & Acesta este un studiu   retrospectiv la pacienţi cu politraum sever, cu fracturi ale căilor femurale,   internaţi în unitatea de terapie intensivă a Spitalului clinic de urgenţă din   Bucureşti şi trataţi din punct de vedere ortopedic, fie conform principiilor   de control al deteriorării (DCO), fie conform principiilor de \textbf{îngrijire totală precoce   (ETC)}.   \\ \hline
Using patients files we   recorded the following data: 30 day mortality, development of acute   respiratory distress syndrome (ARDS) and MODS, local infectious complications   (LIC), intensive care unit length of stay (ICU LOS), days of mechanical   ventilation (MV), units of red blood cells units/48 h (RBC).                   & Utilizând dosarele   pacienţilor, am înregistrat următoarele date: mortalitate cu durata de 30   zile, apariţia sindromului de detresă respiratorie acută (\textbf{SRA}) şi MODS, complicaţii   infecţioase locale (LIC), durata de şedere la unitatea de terapie intensivă   (ICU LOS), zile de ventilaţie mecanică (MV), unităţi de celule roşii în   sânge/48 ore (RBC).       \\ \hline
We decided to analyze   results in three groups - DCO group with shock on admission, DCO group   without shock and ETC group.                                                                                                                                                                                                            & Am hotărât să analizăm   rezultatele în trei grupuri - grupul cu \textbf{COD} cu șoc la admitere, grupul cu COD   fără șoc și grupul cu ETC.                                                                                                                                                                                                                                      \\ \hline
In the other two groups   (DCO without shock and ETC) all outcomes were similar.                                                                                                                                                                                                                                                         & În celelalte două grupuri (\textbf{COD}   fără şoc şi ETC), toate rezultatele au fost similare.                                                                                                                                                                                                                                                                                   \\ \hline
In patients who are not   in a very severe condition (shock), the choice for femoral shaft   stabilization by intramedullary nailing represents a safe option.                                                                                                                                                                           & La pacienţii care nu   sunt într-o afecţiune foarte severă (şoc), opţiunea stabilizării căilor   femurale prin \textbf{nailing   intramedullar} reprezintă o opţiune sigură.                                                                                                                                                                                                      \\ \hline
The biochemical   processes of bioproduction of free radicals (FR) are significantly increasing   in polytrauma patients.                                                                                                                                                                                                                & Procesele biochimice de   bioproducţie a radicalilor liberi (\textbf{RF}) cresc semnificativ la pacienţii cu   politrauma.                                                                                                                                                                                                                                                        \\ \hline
Decreased plasma   concentrations of antioxidants, correlated with a disturbance of the redox   balance are responsible for the installation of the phenomenon called   oxidative stress (OS).                                                                                                                                           & Scăderea concentraţiilor   plasmatice de antioxidanti, corelată cu o tulburare a echilibrului redox,   este responsabilă de instalarea fenomenului numit stres oxidativ (\textbf{SSO}).                                                                                                                                                                                           \\ \bottomrule
\end{tabular}
\caption{Fine-tuned MBart annotated examples for the Other error category. The additional errors present in these examples have not been highlighted in this table.}
\label{tab:err2}
\end{table}

\section{Conclusions and Future Work}

We quantified the impact of domain adaptation on MBart in the medical domain for English-Romanian. The fine-tuned MBart outperforms the general model with automatic metrics and produces fewer errors ($\downarrow$ $34.7$\%) related to terminology in the relatively small sample ($75$ segments belonging to $12$ medical article abstracts) annotated by our annotators. While lower numbers of errors were recorded in the \textit{Partial error, Source term copied, Disambiguation issue in target, Incorrect lexical selection}, and \textit{Term drop}, in the three remaining categories the fine-tuned MBart actually contained more errors than general MBart:\textit{ Inflectional error, Reorder error}, and \textit{Other error}.

Of these three categories, the \textit{Inflectional error}, and \textit{Other error} items present in the fine-tuned MBart output we evaluated are related to the Byte pair encoding (BPE) vocabulary. In future work, we plan to extend the BPE vocabulary in MBart \citep{berard_continual_2021} to cope with in-domain terminology, and to quantify the impact of the fine-tuning on other error types present in the MQM Core. Moreover, we noticed further examples of hallucinations, but they were not within the area of terminology translation, and we will leave them as future work, alongside the additional types of errors noticed in the general MBart and fine-tuned MBart outputs, but also in the reference translations, which were by no means error-free.

More generally, it is essential to raise the awareness of machine translation post-editors, as well as clients, regarding how these error categories are still manifested in MT output even after fine-tuning. NMT output errors remain difficult to identify due to the apparent fluency of the output, and even subject-matter experts can miss some of them. The alert revision and correction of MT output (which has been misleadingly called “postediting” for the past $60$ years \citep{pierce_language_1966} as if it were a monolingual task, not a bilingual one) carries important risks in some settings if assigned to only one person working under high time pressure and using the same text-based revision environments created in the 1990s to accommodate translation memories.

\section*{Acknowledgments}
The GPU used for this research was sponsored by the Google Cloud Research Credits Program.

\bibliographystyle{abbrvnat}

\bibliography{references}

\end{document}